# Using a Diathesis Model for Semantic Parsing

Jordi Atserias, Irene Castellón*, Montse Civit and German Rigau

Departament de Llenguatges i Sistemes Informàtics. Universitat Politècnica de Catalunya
Jordi Girona Salgado, 1-3. Barcelona 08034 Spain
{batalla,civit,g.rigau}@lsi.upc.es
http://www.lsi.upc.es/~acquilex/nlrg.html

* LARELC, Secció de Lingüística General. Universitat de Barcelona
Gran Via de les Corts Catalanes, 585. Barcelona 08007 Spain.
castel@lingua.fil.ub.es
http://www.ub.es/ling/labcat.htm

## Abstract

This paper presents a semantic parsing approach for unrestricted texts. Semantic parsing is one of the major bottlenecks of Natural Language Understanding (NLU) systems and usually requires building expensive resources not easily portable to other domains. Our approach obtains a case-role analysis, in which the semantic roles of the verb are identified. In order to cover all the possible syntactic realisations of a verb, our system combines their argument structure with a set of general semantic labelled diatheses models. Combining them, the system builds a set of syntactic-semantic patterns with their own role-case representation. Once the patterns are build, we use an approximate tree pattern-matching algorithm to identify the most reliable pattern for a sentence. The pattern matching is performed between the syntactic-semantic patterns and the feature-structure tree representing the morphological, syntactical and semantic information of the analysed sentence. For sentences assigned to the correct model, the semantic parsing system we are presenting identifies correctly more than 73% of possible semantic case-roles.

**Keys:**

Semantic Parsing, Semantic Interpretation, Information Extraction, NLU.

## 1 Introduction

Semantic parsing, seen as the mapping from words to semantics to produce a semantic interpretation of sentences (Hirst 87), is a major bottleneck of Natural Language Understanding (NLU) systems. Usually the parser and the semantic analysis are domain dependent. That is, it requires a high cost development of resources which are not easily portable to different domains. Although empirical machine learning methods have proven to be useful in reducing that cost for specific domains (Tou Ng et al. 97), more general approaches are necessary in order to make the systems and the resources more portable to different domains.

Two of the main problems of the production of large-scale semantic analysis are the need to cover all possible semantic realisations of the concepts in a sentence and to produce the same conceptual representation. This task is crucial to obtain correct and complete case-role analysis, in which the semantic roles of the verb such an agent, instrument, etc. are identified.

Our approach obtains a case-role analysis where the semantic roles of the verb are identified. Using general linguistic knowledge, the system automatically builds up the syntactic-semantic patterns of all the possible realisations of the arguments of the verb (each verbal entry includes sub-categorisation structure, semantic roles and selectional restrictions). In order to cover all the possible syntactic realisations of a verb, our system combines their argument structure with a set of general Semantic Labelled Diathesis Models (SLDM). Combining them, the system builds a set of syntactic-semantic patterns with their own role-case representation. Once the patterns are built, we use an approximate tree pattern-matching algorithm to identify the most reliable pattern for a sentence. The pattern matching is performed between the syntactic-semantic patterns and the feature-structure tree representing the morphological, syntactical and semantic information of the analysed sentence. Currently, for sentences assigned to the correct model our system identifies correctly more than 73% of possible semantic case-roles.

Some Information Extraction Systems (such as FASTUS (Appelt 95) or PROTEUS (Grishman 95)) explore similar mechanisms based on "*meta-patterns*" to avoid multiple definition of the same extraction patterns due to the syntactic variations.

Our approach can be seen as a first step of a non-domain specific semantic parser. The system uses a large set of wide coverage tools and resources for Spanish. These tools and resources allow to build a feature-structure (FS) tree analysis of the sentence. This analysis contains morphological, syntactical and semantic information provided by a wide coverage morphological analyser (Carmona et al.'98) and Tagger (Padró 98), a chart parser using a shallow grammar (Castellón et al. 98) and the Spanish EuroWordnet ontology (Farreres et al. 98) (Rodríguez et al. 98).

After this short introduction, Section 2 describes the system architecture and Section 3 explains the preliminary processes performed to produce the complete feature-structure tree analysis. Section 4 is devoted to the construction of the syntactic-semantic patterns by means of the SLDM and the verbal entries. Section 5 focus on the pattern-matching algorithm used to map the feature-structure analysis to the syntactic-semantic patterns. Section 6 describes the experiments carried out and the results achieved. Finally, Section 7 summarises some conclusions and possible further work.

## 2 System Architecture

The Semantic Parsing system we are presenting consists of three different modules:

1. The **Sentence Analyser** performs full syntactic analysis of the sentences. This module is described in next Section.

2. The **Pattern Builder** builds up the sentence models using the verb sub-categorisation and the Semantic Labelled Diathesis Models (SLDM). Section 4 describes this module.

3. The **Tree-Pattern Matcher** chooses the best model for the parsed sentence and builds up the final semantic case-role representation. This module is explained in Section 5.

## 3 The Sentence Analyser

This module produces parsed trees for general domain texts. The Sentence Analyser involves a set of partial steps (i.e. tokenization, morphological analysis and tagging, syntactic parsing and semantic labelling). This process obtains a complete parsed tree for each sentence. Its nodes are lexical features containing lexical information provided by a wide coverage morphological analyser (Carmona et al.'98) and Tagger (Padró 98), and the semantic information from the Spanish Wordnet and the EuroWordnet Top Ontology (Farreres et al. 98). In the example below it is shown the result of the first step of the parsing process: for each word form we obtain the lemma, the disambiguated POS and all the possible EuroWordNet semantic labels.

**Example :**

| | | | |
|---|---|---|---|
| **Con** | con | SPS00 | * |
| **sus** | su | DP3CP00 | * |
| **labios** | labio | NCMP000 | BodyPart\| ... |
| , | , | Fc | * |
| **fue** | ser | VAIS3S0 | Stative\| ... |
| **susurrado** | susurrar | VMPP0SM | Comm.Event\|... |
| **el** | el | TDMS0 | * |
| **secreto** | secreto | NCMS000 | Meaning\| ... |
| . | . | Fp | * |

The final syntactic analysis is produced by a chart parser that uses a wide coverage grammar (Castellón et al. 98) (see figure 1).

## 4 The Pattern Builder

This module completes the Semantic Labelled Diathesis Models (SLDM) using the verb specific information of the sentence to build tree patterns. Any missing information from SLDM is filled with the corresponding information of the verbal entry. If there is no specific information for that verb, the class information is used.

### 4.1 The Semantic Labelled Diathesis Models (SLDM)

We use the theoretical model of diathesis developed in the Pirapides project (Fernández et al. 98). The aim of the Pirapides project is to establish a wide coverage classification for verbs in Spanish and Catalan. The project works in the definition of a theoretical model for verbal entries based on three main components:

- Event structure: Based on (Pustejovsky 95). The model distinguish between simple (event and state) and complex (when there is more than one event/state involved) event structures.

- Meaning Components. They are related to each element the verb subcategorises (including the subject). They are conceived as more abstract than the thematic roles: for instance, the meaning component *iniciador* (starter) includes the roles agent, source and experiencer.
- Diatheses: They are the sintagmatic expression of different semantic oppositions.

### 4.1.1 The diatheses in the Pirapides project

The Pirapides model distinguishes three kinds of semantic oppositions: change of focus, under-specification and aspectual opposition. **Change of focus** appears when there is a change in the point of view between the elements subcategorised by the verb.

i.e. *Los arquitectos construyeron el puente*
 (The architects build the bridge).
 *El puente fue construido por los arquitectos*
 (The bridge was build by the architects)

The **under-specification** appears when a verbal argument is ommited,

i.e. *El profesor dicta ejercicios a los alumnos*
 (The teacher dictate exercises to the pupils)
 *El profesor dicta ejercicios*
 (The teacher dictate exercises)

and finally, the **aspectual opposition** implies a swich from an event to an state.

i.e: *Ana bailó el tango*.
 (Ana danced tango)
 *Ana baila el tango muy bien*
 (Ana is a good dancing tango)

In Pirapides, diatheses are defined as the syntagmatic expressions of a semantic opposition. Diathesis alternations are pairs of structures related to each other by one of those oppositions. From this point of view, the meanings of two sentences expressed with a pair of diatheses don't necessary bear the same meaning.

Taking into account this alternations, verbs can be classified in three main classes according to whether they admit or not those oppositions.

By now, Pirapides has studied and defined the verbal classes of *change of state*, *attitude* and *trajectory*. The trajectory class has been divided in four sub-classes according to two main criteria: whether the verb can express both points of the Trajectory (source and destination) or only one, and whether the verb can express a transfer done independently by the entity or not.

### 4.1.2 Semantic Labelling of the Pirapides diatheses

As described before, SLDM specifies syntactic alternations of verbs (active, passive, anti-causative, etc.) associated with a semantic opposition. Those alternations have been semantically labelled with role-names –*iniciador* (starter), *entidad* (entity), *instrumento* (instrument)- and semantic constraints –*humano* (human), *animado* (animated), *instrumento* (instrument), *causa_natural* (natural_cause)-.

To obtain a full syntactic-semantic pattern of the verb argument structure, SLDM are combined with the syntactic and semantic information of the verb (the preposition that rules the argument, the selectional restrictions and the their possible syntactic realisations).

The elements of the SLDMs include the following information:

- Syntactic categories (and for PPs, preposition).
- Semantic constraints (selectional restriction).
- Morphological information (lemma, word form, gender, number, person).
- Correference with other SLDM elements.
- Agreement with other SLDM elements.
- Optionality of the element.
- Roles (Meaning components)

The diatheses have been classified according to the semantical transitivity of the verb. (semantically intransitive, transitive, transitive using PP).

For instance, Table 1 shows the SLDM for passive voice in transitive verbs. Note that the empty features do not constraint the SLDM.

### 4.2 The Verbal Entries

Verbal entries are described under a syntactic-semantic point of view and are logically organised in a hierarchy of classes. Each verbal entry specifies:

- The semantical transitivity of the verb.
- A list of its arguments/roles with:
  - The syntactic realisation of the role as noun phrase
  - The selectional restriction for the role
  - The preposition when the role appears as a PP.

For instance, table 1 shows the verbal entry for *susurrar* (whisper). As some features are also represented in SLDM both feature structures (FS) can be combined to build a richer model.

## 4.3 Building a Pattern Model

Once a verb is located in the sentence, its verbal entry is combined with the SLDM with the same transitivity to obtain the syntactic-semantic patterns.

Those patterns are built by completing the missing information in the verb-roles from SLDM (the syntactic realisation, selectional restriction, PP preposition) and adding the specific roles of the verb, if any. For instance, table 3 shows how each role of the SLDM corresponding to passive voice (shown in Table 2) is combined with the verbal entry *susurrar* (shown in Table 1) to build up a tree-pattern. **Role entidad:** the verbal entry specifies that the entity role as a NP can be realised as a pronoun (npatons) as a NP (sn) or as a subordinate clause (prop). As neither **vaux** nor **event** appear in the verbal entry no information is added.

**Role iniciador**: This role, in the passive voice, is realised syntactically as a PP with the preposition "*por/de*" (by). If no semantic constraint (selectional restriction) is specified, the selectional restriction corresponding to this role is added from the verbal entry. **Role Meta**: As no preposition is present in the SLDM, this information is taken from the verbal entry "*a/al*" (to). In the same way, the selectional restriction *Human* is added as in the case of *iniciador*.

Once this process is completed, all roles from the verb entry that do not appear in the SLDM and are not *entidad, iniciador, meta* are added as optional. So in the example of *susurrar,* two more roles are added (**entidad.2** and **medio**).

| Model | Trans | | | |
|---|---|---|---|---|
| **Roles** | **NP realisation** | **Preposition for PP** | | **Semantic** |
| Iniciador (starter) | sn/%psubj | | | Human |
| Entidad (entity) | sn/spatons/prop | | | |
| Entidad.2 (entity.2) | sp | De/sobre | | |
| Meta (goal) | sn/npatons | A/al | | Human |
| Medio (instrument) | sp | Con/por/a_través_de | | Instrument |

**Table 1**: Representation of the "*susurrar*" verbal entry.

| Model | Trans | | | | | | |
|---|---|---|---|---|---|---|---|
| **Meaning component** | **Syntax** | **Prep** | **Morph** | **Sem** | **Agreement** | **Co-refer.** | **Optional** |
| Entidad | sn | | | | i | | false |
| Vaux | vser | | | | i | | false |
| Event | vpart | | | | | | false |
| Iniciador | sp | por/de | | | | | true |
| Meta | sp | | | | | | true |

**Table 2**: SLDM representation[1] of the passive voice scheme.

| **Meaning Component** | **Syntax** | **Preposition** | **Morph** | **Semantic** | **Agr.** | **Cor.** | **Opt.** |
|---|---|---|---|---|---|---|---|
| Entidad | *sn/patons/prop* | | | | i | | false |
| Vaux | vser | | | | i | | false |
| Event | vpart | | | | | | false |
| Iniciador | sp | por/de | | *Human* | | | true |
| Meta | sp | a/al | | Human | | | true |
| Entidad.2 | sp | de/sobre | | | | | true |
| *Medio* | *sp* | *con/por/ a_través_de* | | *Instrument* | | | *true* |

**Table 3**: Pattern combination of the SLDM for passive voice and the "*susurrar*" verbal entry.

---

[1] **Syntactic categories**: vser (auxiliar form), vpart (verb in past participle) sn (Noun Phrase), sp (Prepositional Phrase), spatons (pronoun), npatons (pronoun except "se"). **Agr.** : Agreement in number and person. In the SLDM only a chain of agreement appears. **Coref**: Co-reference identifies elements referring to the same entity. In SLDM only one chain co-reference appears.

# 5 The Tree-Pattern Matcher

This module determines which one of the patterns created by the pattern-builder fits the best parse tree. In order to improve coverage, an inexact tree pattern matching algorithm is used. The method we propose is based on the definition of a similarity measure using tree editing operations. We adapt the method proposed by (Tsong-Li et al.'94) to retrieve similar syntactically labelled trees from a Tree Bank for comparing FS parse-trees with tree patterns. The main differences with our approach are:

- The trees are FS trees. The Tree pattern contains expressions such as (*or, not, sub-string*)
- The tree editing operation -**Relabel-** is applied on the features of the structure.
- A tree model operation –**Move-** allowing disordering of the siblings.
- A new constraint in the pattern matching algorithm (the **structural criterion**) to avoid the deletion of some structures than are relevant from a linguistic point of view.
- The addition of some structural heuristics to the cost function.
- Our similarity measure is not a distance measure because we consider the **insert** and **delete** editing operations not always symmetrical. As the **insert** operation adds any kind of information not present in the tree pattern and the **delete** operation could remove relevant verbal arguments, both operations have different cost.

## 5.1 Adapting an approximate tree pattern-matching algorithm to FS parsed-trees

A parsed tree is an ordered tree whose features hold lexical information. The patterns have also FS as nodes whose values can be an expression ("not", "or" and "prefix"), a variable (i.e., to force agreement) or a constant value. We will define the mapping between a tree and a pattern as the function, resulting from the pattern matching, that assigns FS from the tree to another one in the pattern.

A FS of a tree-pattern can be *unified* to another of the parse tree if and only if each feature *unifies* to the same feature of the parsed-tree. Moreover, we impose two restrictions to the tree pattern matching, the *ancestor criteria* and the *structural criteria*.

- **Ancestor-criteria**: The mapped FS of both trees must have the same ancestor relation.

- **Structural-criteria**: This criterion preserves the structures of the parsed-tree through the different levels of the mapping in order to avoid partial structure mappings. For instance, if we are looking for a NP in S, this criterion will avoid the mapping to a NP inside a PP (even if the PP is in S).

As explained before, we define a similarity measure to choose the best (one or more) of all possible matching between the parsed-tree and the pattern. This measure is defined as the minimum cost of all possible sequences of tree editing operations that transform one tree to the other one. The cost of a sequence of operations is the addition of the cost of each operation. We defined the following tree edit operations:

- **Re-label,** this operation changes the value of a feature.
- **Delete,** this operation removes a FS of the pattern tree. There are two kinds of deleting operations, cutting (just this FS) or pruning (the node and all its descendants).
- **Move**, this operation changes the order of siblings.
- **Insert,** this operation adds a FS in the tree.

The cost function assigns a non-negative integer to each editing operation. To make more general the similarity measure between trees, the cost function not only depends on the type of the operation but also on the position of FS in the tree. That is, whether a FS is a leaf (as leaves contain the word forms) or whether the FS does not have an ancestor mapped.

| FS | Syntactic category | Preposition | Morph | Semantic | Agr. with the verb |
|----|--------------------|-------------|-------|----------|--------------------|
| 0 | sp | con | | BodyPart\|.. | true |
| 1 | sn | | | BodyPart\|.. | true |
| 2 | fc | | | | |
| 3 | grup-verbal (verbal group) | | | Stative\|Com | |
| 4 | vser | | | Stative\|.. | |
| 5 | vpart | | | Com.Event | |
| 6 | sn | | | Meaning\|.. | true |

**Table 4**: Main FSs of the parse-tree involved in the matching process

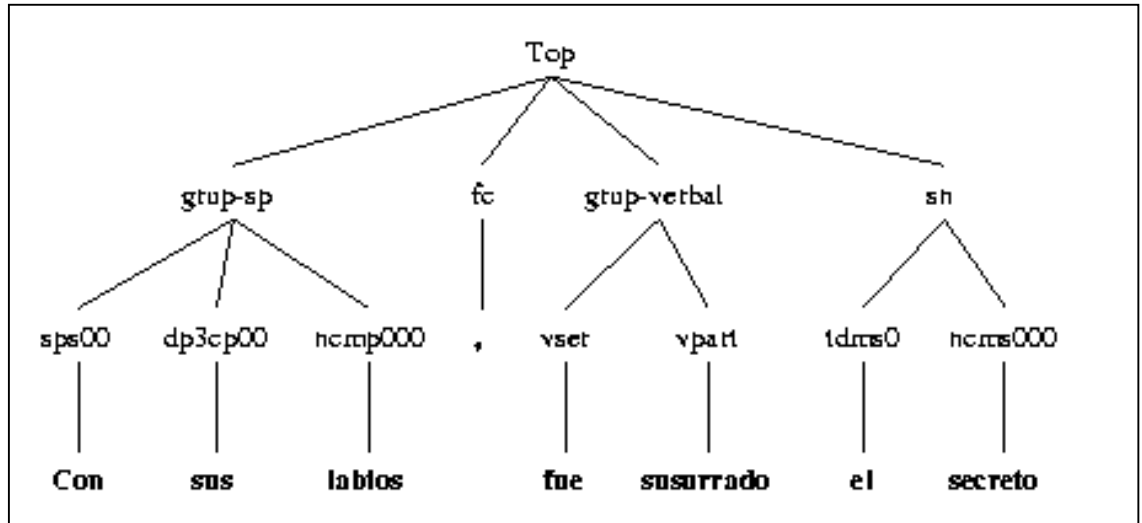

**Fig 1**: Parse tree: *"Con sus labios , fue susurrado el secreto"*

| Meaning Component | Lexical groups |
|-------------------|----------------|
| Entidad | El secreto |
| Vaux | Fue |
| Event | Susurrado |
| Medio | Con sus labios |

**Table 7:** Final result for *"Con sus labios, fue susurrado el secreto"*

For instance, given the sentence *"Con sus labios, fue susurrado el secreto"(literaly, with his/her lips, the secret was whispered),* the application of the syntactic-semantic pattern for the passive voice of the verb "*susurrar*" performs as follows.

The matching process tries to assign each FS in the pattern to another in the parse tree. Figure 1 shows the syntactic structure of the parsed-tree and table 4 its main FSs. The first role in the pattern (shown in table 3), the role *entidad* (an NP in agreement with the verb) could be mapped to the FSs 1 or 6 in the parse-tree. But the structural criterion will prevent from taking the NP 1, *labios,* from inside a PP. FSs with roles *iniciador* and *meta* are deleted as they can not be mapped to any FSs of the parsed-tree. The components, *vser* and *vpart* can only be mapped respectively to FSs 4 and 5 in the parsed-tree. The role *medio* can only be mapped to FS 0 using the move operation and re-labelling the semantic feature from *BodyPart* to *Instrument*. Finally, the algorithm insert the FSs 3,2 and the leaves from the parsed-tree to the pattern as it can not be mapped to any other FS in the pattern.

The resulting similarity between the parsed-tree and the pattern from the SLDM for passive voice of the verb *susurrar*, will be calculated by adding the costs of the two deleting operations, the *move*, the *re-label* and the nine *insert* operations.

## 6 Experiments & Results

As one of the major goals of this work has been to test not only the feasibility of the method but also the linguistic data, we have performed a complete experiment for Spanish using 42 SLDM for the verbs of the **Trajectory Class** developed in **Pirapides** (Morante et al. 98) and ten verbal entries. Eight of these entries belongs to the communication class which is a sub-class of the trajectory class, -*explicar* (to explain), *charlar* (to chat), *decir* (to say), *hablar* (to talk), *murmurar*(to murmur), *susurrar* (to whisper), *discutir* (to discuss), *criticar* (to criticise)-, and two verbs belongs to other classes -*reprender* (to reprimand), *invitar* (to invite)-.

In order to test the generality and soundness of the method we also used a corpus that not only contains verbs of the trajectory class. The sentences of this corpus contain prototypical diathesis alterations.

### 6.1 Semantic Representation

To perform wide-coverage semantic (neither domain specific nor language specific), the 79 semantic labels defined in the preliminary version of the Top Ontology was chosen as a common semantic representation for SLDM, verbal entries and the parse tree. The Top Ontology was developed inside the EuroWordNet project as an ontology for clustering the common base concepts defined for the different languages involved in the project (Vossen et al'97).

### 6.2 Corpus

We have divided the whole corpus in two. The first part has been used for tuning the SLDM models (the tuning corpus) and the second one (the test corpus) for testing the process independently. During the tuning process, we modify the 31 original models adding ten more SLDM produced by splitting the initial models or by taking into account new models. Table 5 summarises some figures of corpus used in the experiment.

|  | **Tuning** | **Test** |
|---|---|---|
| *Sentences* | 257 | 47 |
| *Words* | 1557 | 274 |
| *Com. verbs* | 186 (72%) | 26 (56%) |
| *Other verbs* | 71 (28%) | 21 (44%) |

**Table 5:** Figures of the corpus.

### 6.3 Syntactic Analysis of the corpus

The corpus was processed to obtain a complete parsed tree for each sentence. The nodes of the parse-trees contain lexical features provided by a wide coverage morphological analyser (Carmona et al.'98) and Tagger (Padró 98), and semantic information from the EuroWordnet Top Ontology (Farreres et al. 98) (see section 3).

### 6.4 Results

Although the project is in progress, performing a cycling tuning process on the linguistic data and algorithm, our initial figures seem to be very promising. The current version achieves with a total coverage out of 96%, a precision of 72% in the test corpus SLDM identification task, and a precision out of 73% in the semantic-role identification task. Moreover, due to slightly differences between models, even when an incorrect SLDM has been selected as a solution, the semantic-role identification is correctly performed. Table 6 shows the results in terms of recall and precision focusing on the model.

|  | **Model** | |
|---|---|---|
|  | **Rec.** | **Prec.** |
| **Tuning** | 85% | 88% |
| **Test** | 66% | 72% |

**Table 6**: Recall and precision for the model identification task.

The evaluation criterion for the roles is the exact string equality, for instance the value "*mismo emperador*" (the emperor himself) to fill a role with value "*emperador*" (emperor) will be counted as a missing. Also missing roles o roles that are not in the solution are counted as errors.

### 6.5 Analysing errors

The main sources of misleading information comes from the **Sentence Analyser** module (morphological, semantic and syntactic errors). **Morphological errors** are produced mainly, when the verb is not recognised as a verb. Moreover, errors in POS tagging can produce incorrect syntactic groups during the parsing and furthermore incoherent structures for the SLDM. **Semantic errors** are produced when no semantic labels are found for some words, converting several SLDM to the same patterns and producing an over-generation of solutions. **Syntactic Errors** produced during the parsing process introduce noise in the result. The main causes of syntactic mistakes are produced by noun modifiers, PP-attachment and incorrect identification of sentence boundaries.

## 7 Conclusions & Further Work

This paper has presented a semantic parsing approach for non domain-specific texts. Our approach obtains a case-role analysis, in which the semantic roles of the verb are identified using general domain resources (taggers, shallow parsers and semantic ontologies). In order to cover all the possible syntactic realisations of a verb (or the class model of the verb), our system combines their argument structure with a set of general semantic labelled diathesis models. Combining them, the system builds a set of syntactic-semantic patterns with their own role-case representation. Once the patterns are build, we use an approximate tree pattern-matching algorithm to identify the most reliable pattern for a sentence. The pattern matching is performed between the syntactic-semantic patterns and the FS tree representing the morphological, syntactical and semantic information of the analysed sentence. For sentences assigned to the correct model, the semantic parsing system we are presenting identifies correctly more than 73% of possible semantic case-roles.

Although the results of the experiments are promising for simple sentences, some tuning must be performed on the SLDM to achieve better performance. Improvements on the similarity measure adding statistical information or probabilities to the model could also be tried. Moreover, to design a more general framework, we are planning to formalise the pattern matching and models as a Consistency Labelling Problem (see (Padró 98)) in which different nominal and verbal models can compete for their case-roles assignment.

## Acknowledgements

This research has been partially funded by the Spanish Research Department (ITEM Project TIC96-1243-C03-03 and Spontaneus-Speech Dialogue System In Limited Domains TIC98-423-C06), the Grup de Recerca Consolidat 1997 SGR 00051, and the EU Comission (EuroWordNet LE4003).